\documentclass{article}

\usepackage[utf8]{inputenc}
\usepackage[T1]{fontenc}
\usepackage{microtype}
\usepackage{graphicx}
\usepackage{subfigure}
\usepackage{booktabs} 

\PassOptionsToPackage{hyphens}{url}
\usepackage[accepted]{icml2020}

\usepackage{hyperref}
\usepackage{titlesec}
\usepackage{paralist}
\usepackage[dvipsnames]{xcolor}
\usepackage[font=small]{caption}
\usepackage[acronym,smallcaps,nowarn,section,nogroupskip,nonumberlist]{glossaries}
\usepackage{todonotes}

\hypersetup{%
  colorlinks,
  linkcolor={blue!40!black},
  citecolor={blue!40!black},
  urlcolor={blue!40!black}
}

\glsdisablehyper{}
\newacronym{ABC}{ABC}{approximate {B}ayesian computation}
\newacronym{IS}{IS}{importance sampling}
\newacronym{MH}{MH}{{M}etropolis--{H}astings}
\newacronym{MCMC}{MCMC}{{M}arkov-chain {M}onte {C}arlo}
\newacronym{HMC}{HMC}{{H}amiltonian {M}onte {C}arlo}
\newacronym{SMC}{SMC}{sequential {M}onte {C}arlo}
\newacronym{PPL}{PPL}{probabilistic programming language}
\newacronym{IC}{IC}{inference compilation}

\setdefaultleftmargin{2ex}{2ex}{}{}{}{}
\newcommand{\pyprob}{\texttt{\small PyProb}}
\newcommand{\openmalaria}{\texttt{\small OpenMalaria}}
\newcommand{\covidsim}{\texttt{\small CovidSim}}

\icmltitlerunning{Simulation-Based Inference for Global Health Decisions}

\begin{document}

\twocolumn[
\icmltitle{Simulation-Based Inference for Global Health Decisions}



\icmlsetsymbol{equal}{*}


\begin{icmlauthorlist}
\icmlauthor{Christian Schroeder de Witt}{ox}
\icmlauthor{Bradley Gram-Hansen}{ox}
\icmlauthor{Nantas Nardelli}{ox}
\\\icmlauthor{Andrew Gambardella}{ox}
\icmlauthor{Rob Zinkov}{ox}
\icmlauthor{Puneet Dokania}{ox}
\icmlauthor{N. Siddharth}{ox}
\\\icmlauthor{Ana Belen Espinosa-Gonzalez}{im}
\icmlauthor{Ara Darzi}{im}
\icmlauthor{Philip Torr}{ox}
\icmlauthor{Atılım Güneş Baydin}{ox}
\end{icmlauthorlist}


\icmlaffiliation{ox}{Department of Engineering Science, University of Oxford, UK}
\icmlaffiliation{im}{Department of Surgery and Cancer, Imperial College London, UK}

\icmlcorrespondingauthor{Christian Schroeder de Witt}{cs@robots.ox.ac.uk}


\vskip 0.3in
]



\printAffiliationsAndNotice{}  

\begin{abstract}
The COVID-19 pandemic has highlighted the importance of in-silico epidemiological modelling in predicting the dynamics of infectious diseases to inform health policy and decision makers about suitable prevention and containment strategies.
Work in this setting involves solving challenging inference and control problems in individual-based models of ever increasing complexity.
Here we discuss recent breakthroughs in machine learning, specifically in simulation-based inference, and explore its potential as a novel venue for model calibration to support the design and evaluation of public health interventions.
To further stimulate research, we are developing software interfaces that turn two cornerstone COVID-19 and malaria epidemiology models (\covidsim\footnote{\scriptsize\url{https://github.com/mrc-ide/covid-sim/}} and \openmalaria\footnote{\scriptsize\url{https://github.com/SwissTPH/openmalaria}}) into probabilistic programs, enabling efficient interpretable Bayesian inference within those simulators.
\end{abstract}

\section{Introduction}

Machine learning has a growing role in increasing health service access and efficiency, particularly in resource-constrained settings, making it a valuable tool for the global health community \cite{panch2018artificial,Wahle000798}.
Moreover, the COVID-19 pandemic \cite{who_coronavirus_2020} has underlined the importance of epidemiological modelling and computer simulation in informing the design and implementation of public health interventions at an unprecedented scale \cite{eubank_commentary_2020}.
For many endemic diseases (e.g., malaria), in-silico optimisation of multi-modal intervention portfolios---from mass vaccination to bed nets---is well established \cite{smith_malaria_2017}.
Analogous modelling for COVID-19 interventions, including social distancing \cite{ferguson_strategies_2006}, is mostly unexplored, yet subject to intense public interest \cite{Mahasem1574}.

The adoption of health informatics in worldwide health systems (e.g., OpenMRS \cite{mamlin2006cooking}, mHealth \cite{adibi2015mobile}) enables access to abundant patient-level and aggregated health data \citep{Wahle000798}.
This is fomenting the development of comprehensive modelling and simulation to support the design of health interventions and policies, and to guide decision-making in a variety of health system domains \citep{fone2003systematic,swanson2012rethinking}.
For example, simulations have provided valuable insight to deal with public health problems such as tobacco consumption in New Zealand \cite{tobias2010application}, and diabetes and obesity in the US \citep{zhang2014impact}. They have been used to explore policy options such as those in maternal and antenatal care in Uganda \cite{semwanga2016applying}, and applied to evaluate health reform scenarios such as predicting changes in access to primary care services in Portugal \cite{fialho2011using}. Their applicability in informing the design of cancer screening programmes has been also discussed \cite{ronco2017causal,getaneh2019role}.
Recently, simulations have informed the response to the COVID-19 outbreak \cite{ferguson2020report}.

The process of informing health interventions and policies through simulations generally involves two steps:
\begin{compactdesc}
\item[Model calibration] The extent to which a simulator can reliably inform real-world prediction and planning is bounded by both model discrepancy \cite{brynjarsdottir2014learning} and how well the model has been calibrated to empirical data \cite{andrianakis_bayesian_2015}.
\item[Optimising decision-making] Identifying optimal multimodal intervention strategies and corresponding risks and uncertainties requires searching through potentially vast parameter spaces, which, due to the computational cost of running large simulators (e.g., in some epidemiological studies), usually cannot be exhaustively evaluated \cite{smith_ensemble_2012}.
\end{compactdesc}

Despite their fundamental importance, model discrepancy and calibration of public-health simulators are frequently only informally addressed, or left undocumented \cite{stout2009calibration,punyacharoensin2011mathematical}.
This may be partially explained by the fact that, while numerous methods for formal sensitivity and uncertainty analysis exist \cite{kennedy_bayesian_2001}, they in general do not scale to complex simulators with more than a few dozen parameters \cite{oakley2004probabilistic}.
Similarly, evidence-based decision-making is usually optimised by comparing outcomes on a small number of hand-crafted scenarios and intervention strategies \cite{smith_ensemble_2012}.

\begin{figure*}
\centering
\includegraphics[trim=8mm 0 8mm 0,clip,width=0.994\linewidth]{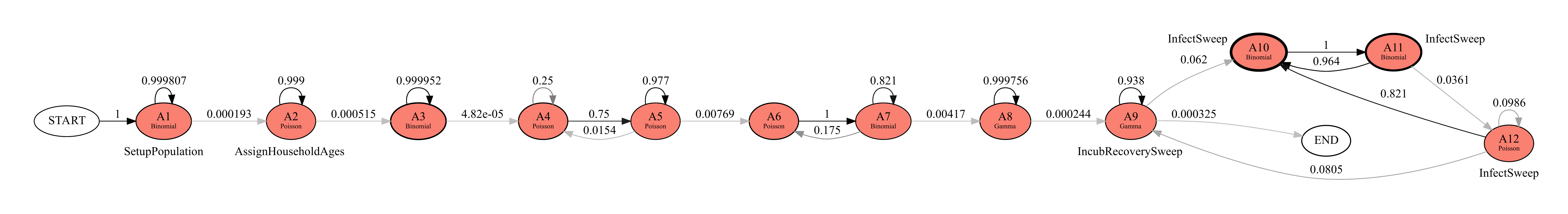}
\caption{
  Latent probabilistic structure uncovered using \pyprob\ from the Imperial College \covidsim\ simulator run on Malta, demonstrating the first step in working with this simulator as a probabilistic program.
  Uniform distributions are omitted for simplicity.}
\label{FigureCovidSim}
\end{figure*}

\section{Epidemiology simulations and inference}

Among the simplest mathematical epidemiology models are deterministic \emph{compartmental models} that partition individuals in a population based on different stages of the disease
\footnote{E.g., susceptible--infectious--recovered or SIR}
\cite{kermack1927contribution,allen1994some,brauer2008compartmental}.
Advances in model construction, computing power, and novel insights into medical and socio-economic aspects have since stimulated the introduction of stochastic \emph{individual-based models}
\footnote{Also referred to as \emph{agent-based model} or \emph{multi-agent system}.}
\cite{railsback2019agent} to public health applications.
These relatively complex and highly-parametrised models are implemented as simulator software that allow studying the global effects of self-organization and emergent properties arising from individual interactions at the local level.

In general, inference in individual-based simulators is usually doubly intractable, as both simulator likelihood and evidence cannot be evaluated efficiently. Likelihood-free methods, including \gls{ABC}~\cite{beaumont2002approximate} have been proposed \cite{andrianakis_bayesian_2015, csillery2010approximate}, but suffer from exponential scaling of inference with data dimension, requiring domain experts to define low-dimensional summary statistics, which ultimately determine quality of inference.

Recent advances in machine learning have led to a new family of promising approaches to simulation-based inference \citep[see][for an overview]{cranmer_frontier_2020}.
In particular, we argue that probabilistic programming \citep[see][for an introduction]{van2018introduction} has a unique potential to standardise and automate model calibration and decision-making in individual-based simulators.

Probabilistic programming allows one to express probability models using computer code and perform statistical inference over the inputs and latent variables of the program, conditioned on data observations (or constraints). This is achieved by using special-purpose probabilistic programming languages (PPLs) \cite{goodman2012church,salvatier2016probabilistic,carpenter2017stan,tran2018simple,bingham2019pyro}, which augment a host language with features to express probabilities and Bayesian conditioning. PPLs separate model specification from inference, allowing flexible selection of appropriate \emph{inference engines} (e.g., \gls{IS} \cite{doucet2009tutorial}, \gls{MCMC} \cite{brooks2011handbook}). 



Recent work made it possible to use pre-existing stochastic simulators as probabilistic programs, with minimal code modification to capture and redirect random number draws \citep{baydin_efficient_2019}, scaling up to very large simulators \citep{baydin_etalumis_2019}, and particularly relevant to this work, with application to individual-based epidemiology simulators \citep{gram-hansen_hijacking_2019}.
Within such a framework, one could, for instance, condition on desired health outcomes (e.g., ICU capacity not being exceeded in a pandemic), and derive detailed posterior distributions over \emph{all} interactions defined by the simulator \cite{wood_planning_2020}, providing insights on interventions effecting a desired outcome---with proper uncertainty quantification at all stages.

To further enhance the applicability of simulation-based inference in this domain we highlight several opportunities for further method development.
Automated amortisation by surrogate methods \cite{gram-hansen_efficient_2019,naderiparizi-2019-amortized,munk_deep_2019}, which aim to automatically identify and replace compute-intensive parts of a simulator through less expensive emulators, could be guided by the causal structure inherent to a simulator (Figure~\ref{FigureCovidSim}), such as many repeated, structurally identical stochastic time steps or multi-agent interactions that might be amenable to mean-field approximations \cite{yang_mean_2018}.
In addition, pre-existing simulators could be turned into differentiable programs \cite{baydin-2018-ad-machinelearning} through automated source-to-source transformations, thus allowing for the use of gradient-based optimisation and inference methods, including Hamiltonian Monte Carlo \cite{neal_mcmc_2012}.
Last but not least, the unified interface specification afforded by probabilistic programming could allow simulators to also become amenable to other techniques from simulation-based inference and control, including dynamic programming and reinforcement learning \cite{levine_reinforcement_2018, van2016black, igl2019multitask}.


To foster the development of a new standardised approach to model calibration and evidence-based decision-making in public health, we are working on instrumenting the existing \covidsim\ \cite{ferguson2020report} and \openmalaria\ \cite{smith2006mathematical} simulators with a probabilistic programming interface through the \pyprob\ library.\footnote{\scriptsize\url{https://github.com/pyprob/pyprob}}
We will publicly release our code to provide out-of-the-box probabilistic programming inference over public health scenarios of interest in these two domains.

We expect the mentioned techniques to play a role in dealing with communicable (infectious) diseases, which already entailed a significant burden for health systems in developing countries \cite{murray2014global} before the worldwide impact of COVID-19.
However, they can also be applicable to non-communicable diseases, such as diabetes and cancer, which are recognised major causes of morbidity and mortality worldwide \cite{ansah2019systems,kruk2018high}.
This will add to the already identified potential of machine learning in health policy \cite{ashrafian2018transforming} and improving health access, emphasising its value for global health in the efforts to achieve universal health coverage and sustainable development goals \cite{panch2018artificial,Wahle000798}.

\newpage
\bibliography{references}

\end{document}